\documentclass[]{llncs}
\usepackage{graphicx}
\usepackage[utf8]{inputenc}
\usepackage{makecell} 

\begin{document}
\title{Automatic Sexism Detection \\ with Multilingual Transformer Models
\subtitle{AIT\_FHSTP@EXIST2021}}

%
%
\author{
Mina Schütz \inst{1} \and
Jaqueline Boeck \inst{2} \and
Daria Liakhovets \inst{1} \and
Djordje Slijepčević \inst{2} \and
Armin Kirchknopf \inst{2} \and
Manuel Hecht \inst{2} \and
Johannes Bogensperger \inst{1} \and
Sven Schlarb \inst{1} \and
Alexander Schindler \inst{1} \and
Matthias Zeppelzauer \inst{2} 
}
\authorrunning{Schütz, Boeck, Liakhovets, Slijepčević, Kirchknopf, Hecht, et al.}
%
\institute{Austrian Institute of Technology GmbH, 1210 Vienna, Austria\\
\email{\{mina.schuetz, daria.liakhovets.fl, johannes.bogensperger, sven.schlarb, alexander.schindler\}@ait.ac.at}\\
\and
St. Pölten University of Applied Sciences, 3100 St. Pölten, Austria\\
\texttt{jaquelineboeck1@gmx.at, manuelhecht8@gmail.com,} \\
\email{\{djordje.slijepcevic, armin.kirchknopf,
 matthias.zeppelzauer\}@fhstp.ac.at}}

{\let\thefootnote\relax\footnotetext{\textit{IberLEF 2021, September 2021, Málaga, Spain.}\\Copyright \textcopyright\ 2021 for this paper by its authors. Use permitted under Creative Commons License Attribution 4.0 International (CC BY 4.0).}}

\maketitle              
\begin{abstract}
Sexism has become an increasingly significant problem on social networks in recent years. The first shared task on sEXism Identification in Social neTworks (EXIST) at IberLEF 2021 is an international competition in the field of Natural Language Processing (NLP) with the aim to automatically identify sexism in social media content by applying machine learning methods. Thereby sexism detection is formulated as a coarse (binary) classification problem and a fine-grained classification task that distinguishes multiple types of sexist content (e.g., dominance, stereotyping, and  objectification). This paper presents the contribution of the AIT\_FHSTP team at the EXIST2021 benchmark for both tasks. To solve the task,s we applied two multilingual transformer models, one based on multilingual BERT and one based on XLM-R. Our approach uses two different strategies to adapt the transformers to the detection of sexist content: first, unsupervised pre-training with additional data and second, supervised fine-tuning with additional and augmented data. For both tasks our best model is XLM-R with unsupervised pre-training on the EXIST data and additional datasets and fine-tuning on the provided dataset. The best run for the binary classification (task~1) achieves a macro F1-score of 0.7752 and scores $5^{th}$ rank in the benchmark; for the multiclass classification (task~2) our best submission scores $6^{th}$ rank with a macro F1-score of 0.5589.


\end{abstract}
\section{Introduction}
Discriminatory views against women are a common occurrence in the online environment. It's detection is challenging since sexism and misogyny may appear in different forms. The first shared task on sEXism Identification in Social neTworks (EXIST) at IberLEF 2021~\cite{EXIST2021,iberlef21} represents a systematic benchmark that attempts to tackle this challenge via machine learning and natural language understanding (NLU). The benchmark covers a wide spectrum of sexist content and aims to differentiate different types of sexist content. This paper presents our contribution to the benchmark, describes our overall approach, the methods and models applied and summarises the obtained results. We  summarise our results for both tasks: the binary sexism identification task (task~1) and the sexism categorization task (task~2). The EXIST benchmark incorporates English and Spanish content from Twitter and Gab which we account for by multilingual modeling.  The peculiarity and contribution of our approach is the use of comprehensive data augmentation and the integration of external (unlabeled) data to make the classification models more robust.  

Our paper is structured as the following: Section \ref{sec:method} describes our methodological approach, describing the employed datasets and models. Our experimental setup will be explained in Section \ref{sec:setup} of this paper, followed by a documentation of the results (Section \ref{sec:results}) and discussion and final conclusion (Section \ref{sec:conclusions}).

\section{Methodological Approach}
\label{sec:method}

A core challenge of the EXIST benchmark is the rather small size of the provided dataset (approx. 7000 training instances). This rather small size makes the robust training of complex NLP methods like transformers difficult. For this reason, we approach the challenge with different transfer learning strategies. As a basis for modeling the textual content, we apply two pre-trained multilingual transformer models: mBERT~\cite{turc2019well} and XLM-R~\cite{XLM-R}. To adapt these general-purpose models to the task of sexism identification and categorization, we propose different data augmentation strategies and extend the dataset with similar content from other datasets. The additional data is used to pre-train and/or fine-tune the transformer models, where in our terminology pre-training refers to unsupervised pre-training and fine-tuning refers to supervised tuning of the classification layers. Our main contribution is the investigation of the following training strategies.

\begin{itemize}
\setlength\itemsep{1em}
\item \textbf{Pre-Training Strategy:} Massively parametrised models such as transformers tend to overfit on small datasets~\cite{schindlercomparing}. To overcome this issue pre-trained models are applied. In our experiments we evaluate different variants of pre-trained transformers and further pre-train them in an unsupervised fashion on semantically related datasets.

\item  \textbf{Fine-Tuning Strategy:} When using pre-trained models, it is necessary to adapt the models to the underlying task. For this purpose, either all layers or only the upper layers of the model are fine-tuned to the task-specific data. Our aim here is to make the higher level feature representations in the model sensitive to the specific task.

\item  \textbf{Fusion Strategy:} As a third strategy we fuse predictions of the best models obtained by the previous two strategies to achieve a prediction.

\end{itemize}
A more detailed description of the implementation of these two strategies is provided in Section~\ref{sec:setup} and the results obtained with these approaches are presented in Section~\ref{sec:results}.

\subsection{EXIST Data}

The challenge contribution is based on the \textit{EXIST2021} dataset which was provided by the EXIST2021 challenge~\cite{EXIST2021}. The dataset contains 6977 training instances in English and Spanish. In total there are 3426 English and 3541 Spanish social media postings from Twitter and Gap. The test set contains 4368 instances, split into 2208 English and 2160 Spanish postings from mentioned sources. They are annotated in a binary fashion (task~1) as either \textit{sexist} or \textit{non-sexist}; and in a more fine-grained categorization (task~2) as: \textit{ideological-inequality}, \textit{objectification}, \textit{stereotyping-dominance}, \textit{misogyny-non-sexual-violence}, \textit{sexual-violence}, \textit{non-sexist}. 

We evaluated the influence of different pre-processing steps on the EXIST dataset (for both languages) covering filtering and normalization of varying intensities:
\begin{itemize}
\item Removing only hashtags: e.g., to avoid over-fitting on specific hashtags.
\item Removing only punctuation
\item Removing mentions, hashtags, and links
\item Removing mentions, hashtags, links, digits, punctuation, and non-ASCII symbols
\end{itemize}
Based on related work on sexism detection \cite{MeTwo} and hate speech detection with transformer models \cite{hate1}, we decided to test different pre-processing pipelines for both languages. Also, corresponding approaches have shown promising results in detecting disinformation with transformer models and using various pre-processing pipelines \cite{fnd}.
Of all pre-processing steps, the last pipeline had the best fine-tuning results for the multilingual approach. Deleting punctuation and non-ASCII symbols seems to have a higher influence on fine-tuning transformer models, when we add Spanish data. Further pre-processing steps such as stopword-removal, stemming or lemmatising were omitted since they are not required by the applied contextualised transformer models or would decrease their performance. For tokenisation the models' built-in tokenisers were used.

\subsection{External Data}

Data augmentation is one of the two strategies being pursued with our challenge contribution. In addition to the EXIST dataset provided by the organisers we pre-train different models on additional datasets which are semantically related to the EXIST dataset. The intention is to learn additional patterns from semantically similar or aligned tasks and to transfer them onto the EXIST tasks.
We conducted experiments using two additional datasets - specifically the  MeTwo~\cite{MeTwo} and HatEval2019~\cite{HatEval} dataset. In our final submissions those datasets were used to pre-train/fine-tune our models.

\begin{itemize}
\setlength\itemsep{1em}
\item \textbf{MeTwo:} is a Spanish dataset which consists of 3600 tweets to detect sexist innuendo, behaviors and expressions. The labels of the tweets are: SEXIST, NON\_SEXIST and DOUBTFUL. The original dataset consists of tweet-IDs labeled as "status\_id" and the associated label for the category. Content and metadata of the corresponding tweets was provided by the creator of the dataset upon request.

\item \textbf{HatEval2019:} is a dataset which can be used for detecting hate speech against women and immigrants. It is composed of 13000 English tweets and 6000 Spanish ones. From a total of 19600 tweets, 9091 have a negative relation towards immigrants and 10509 against women. Furthermore, the tweets are divided into 3 categories: 

\begin{itemize}
    \setlength\itemsep{0.1em}
    \item Hate Speech (HS): Binary value that indicates if hate speech against women or immigrants occurs in the tweet or not. 
    \item Target Range (TR): If hate speech occurs in the tweet, the target range specifies whether it targets a generic group of people or a specific individual.
    \item Aggressiveness (AG): If hate speech occurs in the tweet, additional information is provided whether this is aggressive or not.
\end{itemize}

\end{itemize}
We augmented the EXIST and the additional datasets by translating each post into the respective other language (i.e., from English to Spanish and vice versa).  Due to this procedure, an English and a Spanish version of each dataset was created. 
The online tool Google Translator was used for this purpose. 

\subsection{Models}
To model the textual data, we employed two different transformers~\cite{VSPU17}: multilingual BERT (mBERT) and XLM-RoBERTa (XLM-R).

\begin{itemize}
\setlength\itemsep{1em}
\item \textbf{mBERT} is based on the original BERT (Bidirectional Encoder Representations from transformers) model~\cite{DCLT19}. Unlike the original transformer architecture, BERT only consists of an encoder and is pre-trained on a large dataset containing content from Wikipedia and the BookCorpus. Pre-training the model can be done with two methods, by capturing a sentence in a bidirectional way with the attention mechanism, i.e., Masked Language Modelling and Next Sentence Prediction. However, BERT is only a monolingual model. Thus, we employ mBERT, which is trained on Wikipedia content in 100 languages~ and thus allows for multilingual modeling \cite{bertml}. 
 
\item \textbf{XLM-R} is a multilingual model trained on 100 languages, similar to mBERT. Unlike the latter, XLM-R is not trained on Wikipedia data but on monolingual CommonCrawl data. The model shows improved cross-lingual language understanding in the results shown in the original paper \cite{XLM-R}. It even outperforms mBERT on several standard NLP benchmark tasks \cite{XLM-R}. The model architecture itself is a combination of two transformer models: XLM~\cite{XLM} and RoBERTa~\cite{LOGD19}. The latter is a monolingual optimised version of the original BERT model and does not support the Next Sentence Prediction pre-training variant in order to achieve a better performance than the basic BERT model. In contrast to the basic XLM model, XLM-R is able to recognise the language in the content by itself on the basis of the specified input IDs \cite{hugroberta}.

\end{itemize}

\section{Experimental Setup}
\label{sec:setup}

\begin{figure}[ht]
\includegraphics[width=0.9\textwidth]{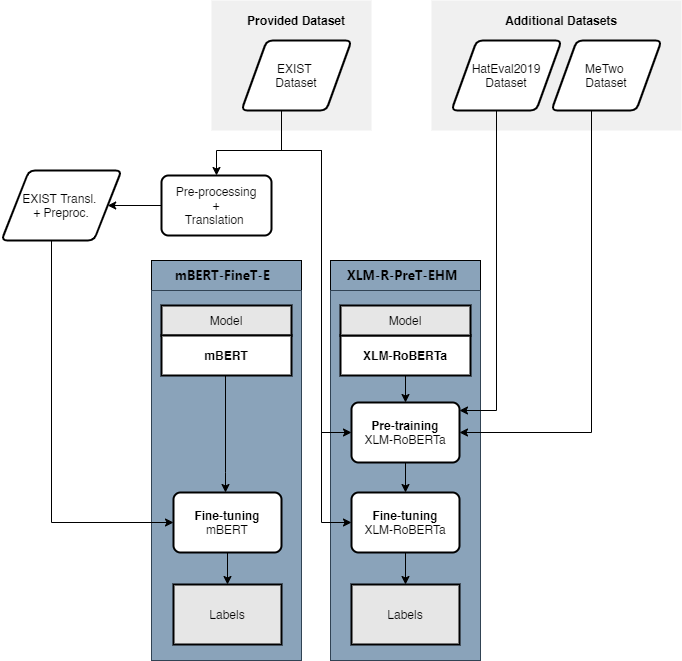}
\caption{Overview of our experimental setup, including two investigated training strategies, i.e., unsupervised pre-training and supervised fine-tuning, and the datasets that are utilised.} 
\label{fig1}
\end{figure}

Figure~\ref{fig1} provides a graphical overview of our experimental setup and the different training strategies. The main focus is on the two investigated approaches, i.e., unsupervised pre-training and supervised fine-tuning, and the datasets that are utilised.

To evaluate the different variants of pre-processing steps we firstly conducted several initial experiments on the EXIST data with multiple pre-trained transformer models provided by the HuggingFace~\cite{HuggingFace} library, such as: BERT~\cite{DCLT19}, RoBERTa~\cite{LOGD19}, ALBERT~\cite{LCGG19}, XLNet~\cite{YDYC19}, and XLM-R~\cite{XLM-R}. We started with the cased BERT model using only the English content and tested each pre-processing pipeline to find the most suitable setup and hyperparameters for our final detection models. The results of those experiments show, that the best outcomes were obtained with no pre-processing, but only when the  Spanish data is not considered. 
We discovered that the overall best results were gained using only English texts on XLNet (80\% for both validation accuracy and macro F1-score). 
The multilingual model XLM-R obtained significantly less accuracy and F1-score with only Spanish data than the monolingual English approaches (with or without pre-processing). 

In the following, we present the setup of the approaches submitted to the benchmark for evaluation. For calculating the evaluation metrics in the development phase we split the provided EXIST training set into 90\% training and 10\% validation (randomly selected).

\subsection{Unsupervised Pre-Training of XLM-R: XLM-R-PreT-EHM}

For this system we used the already pre-trained XLM-R~\cite{XLM-R} and re-trained the model with additional epochs using the RoBERTa Masked Language Modeling (MLM) task on the original (not pre-processed and not translated) EXIST, HatEval2019 and MeTwo datasets.  We pre-trained the model for 25 epochs on each of the datasets, with a batch size of 16, a learning rate of $5e^{-5}$, and AdamW as an optimiser. 
Then we fine-tuned the resulting model for the text classification task, just using the EXIST training data. However, we fine-tuned the model only for task~2 (multi-class classification) and then obtained the labels for task~1 (binary classification) from the multi-class model predictions. We fine-tuned our model for 3 epochs with a batch size of 8, learning rate of $1e^{-5}$, AdamW as an optimiser, 500 warm-up steps and a weight decay of 0.01.

\subsection{Supervised Fine-Tuning of mBERT: mBERT-FineT-E}
We used an already pre-trained multilingual, uncased BERT model (model size: L=12, H=768, A=12; number of total parameters = 110M)~\cite{turc2019well} and fine-tuned it on the provided EXIST dataset and its translations. Beforehand, the data was pre-processed by removing mentions, hashtags, links, digits, punctuation, and non-ASCII symbols. In a first step, we fine-tuned the pre-trained mBERT using only the provided EXIST dataset. Subsequently, we conducted further experiments using the translated EXIST dataset and the additional datasets (HatEval2019 and MeTwo). We fine-tuned the pre-trained mBERT for all combinations of the datasets with and without translations. The best results in the development phase were achieved using only the EXIST dataset and translations. For both tasks, the proposed mBERT was trained separately using an Adam optimiser with a learning rate of $1e^{-5}$ and an epsilon of $1e^{-8}$. We preset the maximum sequence length for our mBERT to 384 and the batch size to 8. Furthermore, we empirically determined the optimal number of epochs for both tasks, i.e., 6 epochs.


\section{Results}
\label{sec:results}

The validation and test results for both tasks are presented in Table~\ref{tab1}. The last column in  Table~\ref{tab1} lists the ranking of our submissions in the EXIST 2021 benchmark. The top ranked submission in the overall benchmark achieved an accuracy of 78.04\% and a macro-averaged F1-score of 78.02\% for task~1 (team: ``AI-UPV'')  and an accuracy of 65.77\% and a macro-averaged F1-score of 57.87\% for task~2 (team: ``AI-UPV'').

\begin{table}[t]
\centering
\caption{Macro-averaged F1-scores (F1) and classification accuracies (CA) for the  sEXism Identification in Social neTworks (EXIST) task at IberLEF 2021. Abbreviation ``val'' stands for our validation set and ``test'' for the official benchmark test set. The performance measures are expressed in percent (\%).}\label{tab1}
\begin{tabular}{c|c|l|c|c|c|c|c}
Task & Run & Approach & CA (val) & F1 (val) & CA (test) & F1 (test) & Ranking\\
\hline
1 & 1 & mBERT-FineT-E &  79.97 & 79.97 & 71.82 & 71.21 & $36^{th}$\\
1 & 2 & XLM-R-PreT-EHM & 79.94 & 79.92 & \textbf{77.54} & \textbf{77.52} & $5^{th}$ \\
1 & 3 & Late Fusion & ----- & ----- & 76.65 & 76.56 & $10^{th}$ \\
\hline
2 & 1 & mBERT-FineT-E &  68.24 & 59.76 & 60.74 & 51.95 & $29^{th}$\\
2 & 2 & XLM-R-PreT-EHM & 68.48 & 59.40 & \textbf{64.45} & \textbf{55.89} & $6^{th}$\\
2 & 3 & Late Fusion & ----- & ----- & 64.45 & 55.59 & $8^{th}$ \\
\end{tabular}
\end{table}

\subsection{Task 1}

\paragraph{\textbf{Fine-Tuning Strategy:}}

For run~1, we used the mBERT fine-tuned on the (pre-processed) EXIST dataset with translations. In our approach in run~1, mBERT seems to overfit on the training data, as the validation accuracy of 79.97\% is significantly higher than the test accuracy of 71.82\%. 

\paragraph{\textbf{Pre-Training Strategy:}}

In run~2, we aggregated the predictions from the XLM-R approach trained for task~2, where we pre-trained the model on the EXIST, HatEval2019 and MeTwo datasets (without translations). Our approach with pre-training XLM-R in run~2 achieves the best results. These results are closely followed by the late fusion approach. The performance in run~2 (and run~3) is similar for our validation and the test set, which indicates that this approach generalises well. Our run~2 ranked $5^{th}$ overall in the benchmark and performed only 0.52\% less accurate (in terms of classification accuracy) than the overall best submission in the EXIST benchmark. 

We conducted experiments with the original XLM-R that we fine-tuned on the original (non pre-processed) EXIST dataset and the additional datasets (see Table~\ref{tab2}). Interestingly, the model performed significantly better for English content than for Spanish content. Fine-tuning on the additional datasets did not improve the results, but rather made them worse. Comparing the results from the last row in Table~\ref{tab2} with our run~2 for task~1 from Table~\ref{tab1}, we can see that the pre-training yielded an advantage over the fine-tuning. 

\begin{table}[t]
\centering
\caption{Additional experimental results for the original XLM-R fine-tuned on the original (non pre-processed) EXIST data and the additional datasets, respectively. The results are obtained on our validation set (from the pre-processed EXIST dataset). The performance measures are expressed in percent (\%).}\label{tab2}
\begin{tabular}{l|l|c|c}
Approach & Validation & CA (val) & F1 (val) \\
\hline
XLM-R fine-tuned on EXIST (EN) & EXIST (EN) & 73 & 70 \\
XLM-R fine-tuned on EXIST (ES) & EXIST (ES) & 53 & 35 \\
XLM-R fine-tuned on EXIST (EN \& ES) & EXIST (EN \& ES) & 72 & 68 \\
\hline
\makecell[l]{XLM-R fine-tuned on EXIST (EN \& ES), \\ HatEval2019, and MeTwo} & EXIST (EN \& ES) & 68 & 63 \\
\end{tabular}
\end{table}

\paragraph{\textbf{Fusion Strategy:}}

For run~3, we performed a late fusion. The predictions were determined by calculating the maximum of the the sum of the predicted class-wise probabilities of  run~1, run~2, and an additional mBERT model fine-tuned on the (pre-processed) EXIST and MeTwo dataset (without translations). Our run~3 performed slightly less accurate (in terms of classification accuracy) than run~2 and ranked $10^{th}$ overall in the benchmark.

\subsection{Task 2}

\paragraph{\textbf{Fine-Tuning Strategy:}}

For task~2 and run~1, we again used the mBERT fine-tuned on the (pre-processed) EXIST dataset with translations. The results indicate that mBERT seems to overfit on the training data, as the validation accuracy of 68.24\% is significantly higher than the test accuracy of 59.76\%. 

\paragraph{\textbf{Pre-Training Strategy:}}

In run~2, we applied the XLM-R approach, where we pre-trained the model on the EXIST, HatEval2019 and MeTwo datasets (without translations). For task~2, a similar pattern can be seen in the results as for task~1. Our approach in run~2 achieved the best results of our or submissions and ranked $6^{th}$ in the EXIST Challenge, performing only 1.98\% less accurate (in terms of macro-averaged F1-score) than the overall best submission in the benchmark. 

\paragraph{\textbf{Fusion Strategy:}}

For run~3, we also performed a late fusion in a similar manner as for task~1, but only with the predicted probabilities of run~1 and run~2. The classification accuracy of the late fusion approach in run~3 is identical to the result in run~2. For the macro F1 score of 55.59\%, results show a slight difference compared to run~2.

\section{Discussion \& Conclusion}
\label{sec:conclusions}

In this paper, we described our submission to the EXIST2021 benchmark, which consists of two tasks on the classification of sexist content. In our experiments we found that the unsupervised pre-training strategy of the XLM-R model~\cite{XLM-R} with additional external data is the most promising strategy, leading to an F1-score of 77.52\% in task~1 and  55.89\% in task~2. The fine-tuning strategy of the mBERT model alone using our augmented corpus is outperformed by the former strategy and shows signs of overfitting. In general, the use of additional data (either external datasets or translations) resulted in improvement for both strategies. As a final remark, our experiments reveal that the fine-tuning of the whole model on domain-specific data was more effective compared to the pure re-training of the classification layer only.

\section{Acknowledgements}

This contribution has been funded by the FFG Project ``Defalsif-AI'' (Austrian security research programme KIRAS of the Federal Ministry of Agriculture, Regions and Tourism(BMLRT), grant no. 879670) and the FFG Project ``Big Data Analytics'' (grant no. 866880).

\bibliographystyle{splncs04}
\bibliography{bib}

\end{document}